%% file: paper.tex
\documentclass{article}
\PassOptionsToPackage{numbers, compress}{natbib}
\usepackage[preprint]{neurips_2026}

\usepackage[utf8]{inputenc}
\usepackage[T1]{fontenc}
\usepackage{hyperref}
\usepackage{url}
\usepackage{booktabs}
\usepackage{amsfonts}
\usepackage{amsmath}
\usepackage{microtype}
\usepackage{xcolor}
\usepackage{graphicx}
\usepackage{xspace}
\newcommand{\method}{\textit{MotionPyramid}\xspace}
\newcommand{\name}{\textit{Residual Interfaces}\xspace}

\usepackage{subcaption}

\title{\method: Hierarchical Motion Representation and Residual Interfaces}

\author{%
  Gao Zhu \\
  UC Davis \\
  \And
  Zaishuo Xia \\
  UC Davis \\
  \And
  Yubei Chen \\
  UC Davis \\
}

\begin{document}

\maketitle

\begin{abstract}

\input{chaps/abstract}
\end{abstract}

\input{chaps/introduction}

\input{chaps/method}
\input{chaps/experiments}
\input{chaps/related_work}
\input{chaps/discussion}

\input{chaps/references}

\appendix

\input{chaps/appendix}

\end{document}

%% file: chaps/abstract.tex
Perceptual systems exhibit a natural representational hierarchy, progressing from local primitives such as edges and blobs to higher-level structures such as parts, objects, and faces. We ask whether an analogous hierarchy can be established for motion. In humanoid control, low-level actions specify immediate motor commands, while meaningful behavior is often organized over longer temporal scales, including contacts, gait fragments, balance recovery, reaching motions, and whole-body skills. We introduce \emph{MotionPyramid}, a hierarchical action representation that learns such structure from motion data. Starting from a motion-tracking teacher, \method trains a recursive stack of latent decoders: low-level latents decode to immediate full-body motor commands, while higher-level latents unfold through lower levels into temporally extended motion programs. After pretraining, the hierarchy is frozen and reused by downstream reinforcement learning policies as a family of action interfaces operating at different control resolutions. Experiments show that the learned levels form a motion hierarchy: coarser interfaces improve early learning and motion regularity by constraining exploration to structured motion segments, while finer interfaces preserve feedback control and final task precision. Representation probes further show that the hierarchy supports traversal, interpolation, transition, and qualitative composition, exposing editable control handles across temporal scales. Finally, we introduce \textit{Residual Interfaces}, which allow a downstream policy to maintain coarse, segment level, and frame level residual commands through the frozen hierarchy. Analogous in spirit to residual or skip connections in deep perceptual and language models, this mechanism allows coarse motion programs and fine residual corrections to coexist within one controller. \method demonstrates that motion, like perception, can be organized into a reusable multi-level representation, providing structured abstraction without sacrificing controllability.

%% file: chaps/introduction.tex
\section{Introduction}
\begin{figure*}[t]
\centering
\includegraphics[width=\textwidth]{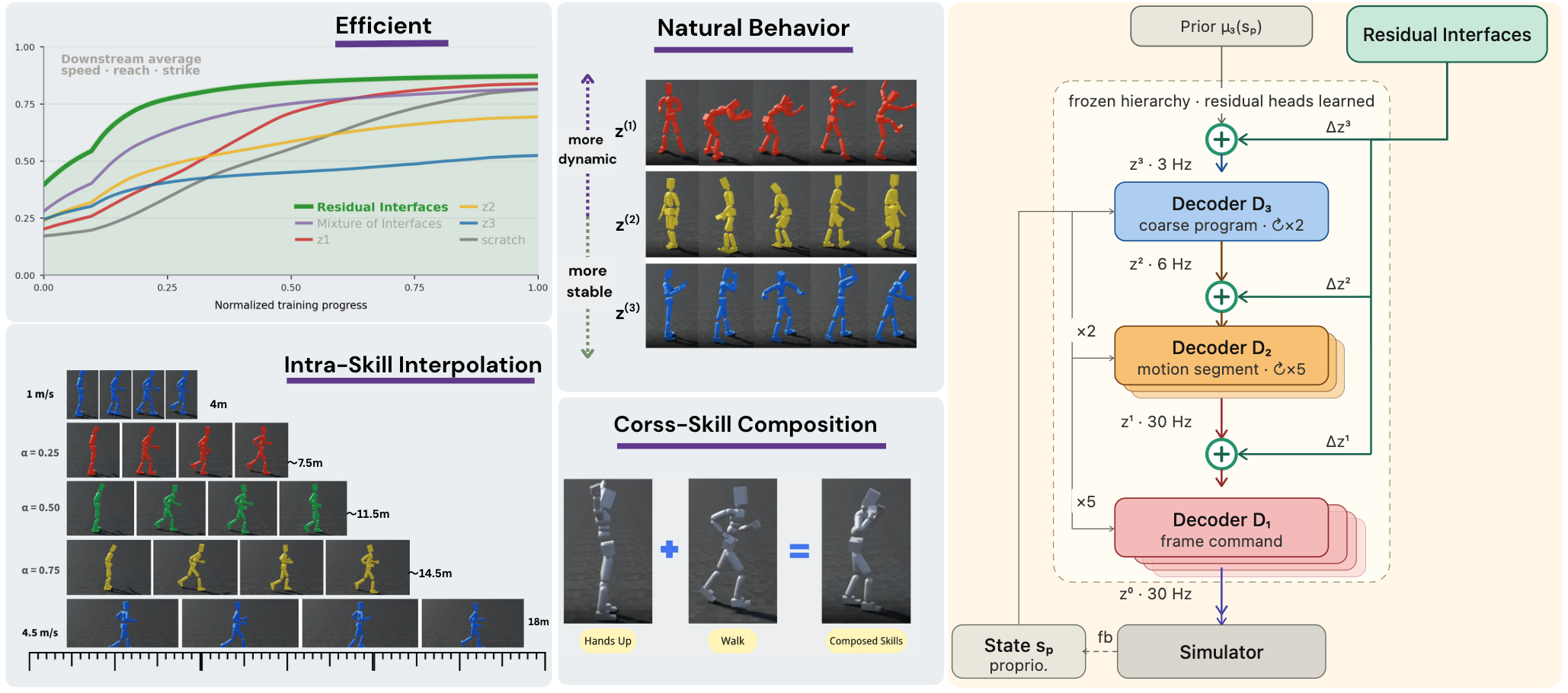}
\caption{
\textbf{Overview of \method}. Left:  representation probes visualize sampling,
interpolation, traversal, and composition through the frozen hierarchy. Right: a hierarchy of reusable action interfaces
recursively unfolds coarse latent decisions into lower level latents and motor
commands. Middle: fixed pyramid levels reveal a tradeoff between learning speed
and final precision across downstream tasks, while \name combines
coarse context, segment refinement, and frame-level correction.
}

\vspace*{-2em}
\label{fig:teaser}
\end{figure*}
Perceptual representation learning provides a familiar example of hierarchy. In vision, early representations often capture local primitives such as oriented edges, blobs, and textures, while deeper representations organize larger structures such as parts, objects, and faces~\citep{zeiler2014visualizing, yosinski2015understanding}. More broadly, analyses of deep vision~\citep{he2016deep} and language models~\citep{vaswani2017attention} suggest that useful information is distributed across multiple depths of a learned representation~\citep{yun2021transformer}. We ask whether motion admits an analogous hierarchy. For humanoid control, this hierarchy is not primarily semantic, but temporal and control-oriented: low levels should preserve immediate feedback, joint coordination, contact handling, and balance correction, while higher levels should represent longer motor programs such as contact patterns, gait fragments, reaching segments, recovery behaviors, and whole-body skills.

The action interface through which a controller acts is a central design choice in humanoid reinforcement learning~\citep{peng2018deepmimic, peng2022ase, luo2024universal}. A simulated humanoid is often controlled by a high-dimensional vector of joint targets at every control step. This interface is expressive, but it gives the policy little prior structure: coordination across joints, contact timing, smooth temporal evolution, and recovery behavior must all be discovered through reward. Classical studies of human and animal motor control have long emphasized that movement is organized through coordinated patterns, feedback correction, and temporally extended motor programs~\citep{bernstein1967coordination,schmidt1975schema,todorov2002optimal,ijspeert2013dynamical,paraschos2013probabilistic}. Computational hierarchical control similarly studies controllers in which higher-level modules direct lower-level controllers through reusable interfaces~\citep{wayne2014hierarchical}. These perspectives suggest that motion should not be represented only as instantaneous motor commands, but as a hierarchy of reusable control abstractions.

Motion data has become a powerful source of structure for physics-based humanoid control. Motion tracking methods can transform reference motion into physically valid control policies, from individual clips to large motion datasets~\citep{peng2018deepmimic,luo2023phc,mahmood2019amass}. Adversarial motion priors and learned latent skills further show that motion data can bias policies toward physically plausible and visually natural behavior during downstream learning \citep{peng2021amp,peng2022ase,tessler2023calm}. Recent work on emergent action representations shows that embeddings learned through multi-task policy training can expose high-level action interfaces useful for interpolation and composition in downstream control \citep{hua2023simple}. Universal humanoid representations further demonstrate that large and diverse motion collections can be distilled into reusable control spaces for many downstream tasks \citep{luo2024universal}. Together, these results establish an important principle: the action space itself can be learned from motion data. This view suggests a path toward motion pretraining, where the reusable pretrained object is not a task-specific policy, but a motion representation that serves as an action interface for future controllers.

A common limitation remains in how learned action spaces are exposed to downstream control. Most methods provide a single action interface, such as raw motor commands, a per-step latent action, or a fixed-horizon skill embedding \citep{merel2018npmp,haarnoja2018latent,yao2022controlvae,won2022physicsvae,zhu2023ncp,peng2022ase,tessler2023calm,luo2024universal,tirinzoni2025metamotivo}. This fixed control resolution couples two competing requirements. Fine interfaces preserve rapid feedback correction, accurate contacts, and precise target alignment, but require frequent decisions in a large or flexible action space. Coarser interfaces make exploration easier by producing coherent motion segments, but can restrict the controller when the task requires precise timing, local recovery, or end-effector accuracy. A single downstream task may require both regimes at different moments.

We introduce \method, a hierarchical motion representation for physics-based humanoid control. \method learns a stack of latent decoders that compress control across body space and time: low-level latents decode to immediate full-body motor commands, while higher-level latents recursively unfold through lower levels into temporally extended motion programs. The hierarchy is pretrained bottom-up from a motion-tracking teacher, frozen, and reused by downstream reinforcement learning policies as a family of residual action interfaces. Rather than producing motor commands directly, a downstream policy predicts residual latents around pretrained motion priors at the selected level. In this sense, \method treats motion pretraining as representation learning for control: the reusable pretrained object is the interface through which future policies act.

The resulting hierarchy exposes complementary control regimes. Higher levels reduce decision frequency and bias exploration toward coherent motion segments, while lower levels preserve local feedback, contact correction, and precise target alignment. We evaluate these levels both as fixed downstream interfaces and as editable representations through sampling, traversal, interpolation, transition, and qualitative composition probes. We introduce two adaptive downstream controllers for using these complementary regimes. Mixture of Interfaces selects one interface level online and predicts a residual latent at that level, while \emph{Residual Interfaces} maintain active coarse, segment level, and frame level residual commands through the frozen hierarchy. This is related to residual connections~\citep{he2016deep} and skip or pyramid connections~\citep{lin2017feature, ronneberger2015u} in deep architectures, while here the access paths operate over temporal control interfaces, allowing coarse motion context and fine feedback correction to coexist within one controller.

This work makes three contributions: 1) We introduce \method, a hierarchical motion representation that pretrains a reusable family of residual action interfaces for downstream humanoid reinforcement learning; 2) We characterize the learned hierarchy through downstream control and representation probes, showing that different levels trade off learning efficiency, motion regularity, editability, and final precision; 3) We introduce two adaptive downstream controllers for the frozen hierarchy: Mixture of Interfaces, which selects one temporal interface online, and \name, a nested residual cascade that maintains active commands at all levels. Residual Interfaces combines slow high level motion context, middle level segment refinement, and frame level correction within one controller.

%% file: chaps/method.tex
\vspace{-1em}
\section{\method}
\vspace{-0.7em}

\method learns a hierarchy of latent action representations for humanoid control. The hierarchy compresses full body motor commands across body space and time, and exposes each level as a reusable action interface for downstream reinforcement learning. The lowest level decodes a compact latent into a motor command at every control step; higher levels decode one latent into a temporally extended sequence of lower level latents that are executed recursively.

We denote simulator control steps by $t$ and the motor command by $z^{(0)}_t \in \mathbb{R}^{69}$. \method learns interfaces
$z^{(3)} \rightarrow z^{(2)} \rightarrow z^{(1)} \rightarrow z^{(0)}$, where higher indexed levels operate at coarser temporal scales and lower indexed levels provide finer control. Let $H_\ell$ be the number of simulator steps covered by one level $\ell$ decision, and let \(M_\ell = H_\ell/H_{\ell-1}\) for $\ell>1$. In our three level instantiation, \(H_1=1\), \(H_2=5\), and \(H_3=10\), so \(M_2=5\) and \(M_3=2\). These correspond to 30 Hz, 6 Hz, and 3 Hz action interfaces, where 30 Hz denotes the decimated policy control rate rather than the internal physics integration rate. We treat these horizons as representative control resolutions for studying the tradeoff between learning efficiency, motion regularity, and final control precision.

\vspace{-0.5em}
\subsection{Teacher Supervision and Base Latent Interface}
\vspace{-0.5em}
\method is pretrained from a full body motion tracking teacher that imitates reference humanoid motion clips and outputs physically valid motor commands. At each control step, the teacher observes the current humanoid state and reference motion state and produces a target motor command $z_t^{(0)^\star}$. These commands supervise the first latent action interface, and the encoded latent rollouts later become targets for higher temporal levels.

The base interface compresses each teacher command into a compact latent action. Given proprioception \(s_t^p\), the decoder \(D_1\) maps a latent \(z_t^{(1)}\) to a motor command, \(z^{(0)}_t = D_1(z_t^{(1)}, s_t^p)\). We also learn a proprioception conditioned diagonal Gaussian prior \(p_1(z_t^{(1)} \mid s_t^p)\). Following common motion prior formulations~\citep{luo2024universal,tessler2024maskedmimic}, an encoder infers \(z_t^{(1)}\) from teacher rollouts and \(D_1\) reconstructs $z_t^{(0)^\star}$. The objective combines reconstruction, KL regularization to the prior, and smoothness. After training, \(D_1\) is frozen and reused as both the lowest level downstream interface and the terminal decoder for higher temporal interfaces. Full objective details are provided in the appendix.
\vspace{-0.5em}
 \subsection{Recursive Temporal Interfaces}
\vspace{-0.5em}
Higher levels of \method perform temporal compression by decoding one
coarse latent into a sequence of lower level latents. For $\ell>1$, a level $\ell$ latent selected at boundary time $b$ unfolds into
$M_\ell$ lower level latents at the corresponding child boundaries:
\[
z^{(\ell-1)}_{b+jH_{\ell-1}}
=
D_\ell\!\left(
z^{(\ell)}_b,\,
s^p_{b+jH_{\ell-1}},\,
\phi^{(\ell)}_j
\right),
j=0,\ldots,M_\ell-1 .
\]
Here $\phi^{(\ell)}_j$ denotes the phase of the child decision inside the parent
temporal block. The generated child latents are then executed recursively by the
lower level interface until \(D_1\) produces motor commands. In our setting, one \(z^{(2)}\) produces five \(z^{(1)}\) latents, while one \(z^{(3)}\) produces two \(z^{(2)}\) latents and covers ten simulator steps through the recursive path \(z^{(3)} \rightarrow z^{(2)} \rightarrow z^{(1)} \rightarrow z^{(0)}\). The unfolding remains closed loop because each decoder conditions on the current proprioceptive state at the child boundary.
\vspace{-0.5em}
\subsection{Closed Loop Distillation}
\vspace{-0.5em}
Temporal interfaces are trained with closed-loop distillation using the same tracking teacher. At each level, the student is rolled out through the current decoders while an encoder observes a short teacher segment and infers the latent action. The decoder then reconstructs the corresponding lower level latent targets.

Let $\bar z^{(1)}_t$ denote the first level target at simulator step $t$, and let $\bar z^{(2)}_b$ denote the second level target at a second level boundary $b$. For $\ell\in\{2,3\}$, the reconstruction target for a level $\ell$ block starting at boundary $b$ is
\[
\bar{\mathbf z}^{(\ell-1)}_{b}
=
\left\{
\bar z^{(\ell-1)}_{b+jH_{\ell-1}}
\right\}_{j=0}^{M_\ell-1}.
\]
Thus \(D_2\) reconstructs a length \(H_2\) sequence of first level latents, while \(D_3\) reconstructs \(M_3\) second level boundary latents rather than a simulator step sequence of length \(H_3\). Each level also learns a proprioception conditioned prior \(p_\ell(z^{(\ell)} \mid s^p)\). The full objective combines latent sequence reconstruction, KL regularization to the prior, and temporal regularization. We train recursively: \(D_1\) is learned first, then frozen while training \(D_2\), and \(D_1,D_2\) are frozen while training \(D_3\).
\vspace{-0.5em}
\subsection{Downstream Control with Fixed Interfaces}
\vspace{-0.5em}
After pretraining, all decoders and priors are frozen. A downstream policy can use any level $\ell$ as its action interface, with the learned prior
\[
p_\ell\!\left(z^{(\ell)} \mid s^p\right)
=
\mathcal N\!\left(
\mu_\ell(s^p),\,
\operatorname{diag}(\sigma_\ell^2(s^p))
\right).
\]
During downstream reinforcement learning, the policy outputs a residual around the prior mean:
\[
\Delta z^{(\ell)}_b \sim \pi^{(\ell)}_\psi(\cdot \mid o_b),
z^{(\ell)}_b
=
\mu_\ell(s^p_b)
+
\Delta z^{(\ell)}_b .
\]
The PPO likelihood is computed over the residual \(\Delta z^{(\ell)}_b\); the prior mean only centers the latent action. The selected latent is unfolded through the frozen pyramid for $H_\ell$ simulator steps. Higher levels reduce policy decision frequency and encourage temporally coherent exploration, while lower levels expose finer feedback control for contacts, target alignment, and rapid recovery.
\vspace{-0.5em}
\subsection{Adaptive Downstream Interfaces}

Fixed interfaces require choosing one temporal scale for an entire downstream task. \method also supports two adaptive controllers over the same frozen hierarchy: Mixture of Interfaces selects one interface level online, while Residual Interfaces keep all levels active through a nested residual cascade.

\paragraph{Mixture of Interfaces.}
Mixture of Interfaces gives the downstream policy a discrete choice among pretrained interfaces. At decision boundary \(b\), the policy samples a level
\[
g_b \sim \pi^g_\psi(\cdot \mid o_b),
\qquad
g_b \in \{1,2,3\},
\]
and then samples a residual latent for the selected level,
\[
\Delta z^{(g_b)}_b
\sim
\pi^{(g_b)}_\psi(\cdot \mid o_b),
\qquad
z^{(g_b)}_b
=
\mu_{g_b}(s^p_b)
+
\Delta z^{(g_b)}_b .
\]
The selected latent is decoded by the corresponding frozen path and executed for
\(H_{g_b}\in\{H_1,H_2,H_3\}\) simulator steps.

The PPO log probability is
\[
\log \pi_\psi
\left(
g_b,\Delta z^{(g_b)}_b \mid o_b
\right)
=
\log \pi^g_\psi(g_b \mid o_b)
+
\log \pi^{(g_b)}_\psi
\left(
\Delta z^{(g_b)}_b \mid o_b
\right).
\]
All likelihood and entropy terms are computed in the native latent dimension \(d_{g_b}\) of the selected interface. In our current instantiation, all learned interfaces use 32-dimensional latent actions, so the likelihood, entropy, PPO ratio, and decoder dispatch operate on the same 32-dimensional residual vector.

Mixture of Interfaces provides an interpretable adaptive controller. Its selected horizon reveals where the policy spends feedback bandwidth, and its average decision frequency is
\[
\bar f_{\mathrm{dec}}
=
f_{\mathrm{sim}}
\frac{N}{\sum_{n=1}^{N} h_n},
\]
where \(h_n=H_{g_n}\) is the duration selected at the \(n\)th decision.

\paragraph{Residual Interfaces.}
Residual Interfaces use a nested residual cascade instead of a discrete selector. At episode reset, all latent commands are refreshed. During rollout, the high level command is refreshed every \(H_3=10\) simulator steps, the middle level command every \(H_2=5\) steps, and the first level command every simulator step.

At a high level boundary \(B\), the policy samples a residual around the frozen high level prior,
\[
\Delta z^{(3)}_B
\sim
\pi^{(3)}_\psi(\cdot \mid o_B),
\qquad
z^{(3)}_B
=
\mu_3(s^p_B)
+
\Delta z^{(3)}_B .
\]
At a middle level boundary \(b=B+jH_2\), where \(j\in\{0,1\}\), the frozen decoder \(D_3\) maps the current high level command into a middle level base command, which is refined by a learned residual,
\[
\begin{aligned}
\Delta z^{(2)}_b
&\sim
\pi^{(2)}_\psi(\cdot \mid o_b),\\
z^{(2)}_b
&=
D_3
\left(
z^{(3)}_B,
s^p_b,
\phi^{(3)}_j
\right)
+
\Delta z^{(2)}_b .
\end{aligned}
\]
At each simulator step \(t=b+i\), where \(i\in\{0,\ldots,H_2-1\}\), the frozen decoder \(D_2\) maps the current middle level command into a first level base latent. The policy then applies a frame level residual before the frozen decoder produces the motor command,
\[
\begin{aligned}
\Delta z^{(1)}_t
&\sim
\pi^{(1)}_\psi(\cdot \mid o_t),\\
z^{(1)}_t
&=
D_2
\left(
z^{(2)}_b,
s^p_t,
\phi^{(2)}_i
\right)
+
\Delta z^{(1)}_t,
\qquad
z^{(0)}_t
&=
D_1
\left(
z^{(1)}_t,
s^p_t
\right).
\end{aligned}
\]

All pretrained decoders and priors remain frozen; only the downstream residual heads, actor, and critic are optimized. Residual commands are held between refresh boundaries, so \(z^{(3)}\) supplies slowly varying context, \(z^{(2)}\) provides segment level refinement, and \(z^{(1)}\) retains frame level correction.

Let \(m^{(3)}_t\), \(m^{(2)}_t\), and \(m^{(1)}_t\) denote refresh masks for the residual heads, with \(m^{(3)}_t=1\) at high level boundaries, \(m^{(2)}_t=1\) at middle level boundaries, and \(m^{(1)}_t=1\) at every simulator step. At episode reset, all masks are set to one. The PPO log probability includes only residuals sampled at step \(t\):
\[
\log \pi_\psi(u_t \mid o_t)
=
\sum_{\ell=1}^{3}
m^{(\ell)}_t
\log
\pi^{(\ell)}_\psi
\left(
\Delta z^{(\ell)}_t
\mid
o_t
\right).
\]
Entropy and PPO ratio terms use the same masks. The critic remains a standard per step value function. Residual Interfaces therefore combine coarse temporal context and fine feedback correction without committing to a single selected interface level.

%% file: chaps/experiments.tex
\vspace{-0.8em}
\section{Experiments}
\vspace{-0.8em}
We evaluate \method from four perspectives. 
First, we study each pyramid level as a fixed downstream action interface and measure the tradeoff between learning efficiency and final task performance. 
Second, we examine motion naturalness through random sampling and latent traversal. 
Third, we probe the learned representation through multiscale motion editing, including interpolation, transition, and composition demos. 
Finally, we analyze adaptive downstream control and study where its gains come from through interface selection and multi-level residual correction. 
Figure~\ref{fig:main_results} summarizes the main findings, Figure~\ref{fig:ablation} compares against 30 Hz latent baselines, and Figure~\ref{fig:selection} analyzes the interface selection behavior of Mixture of Interfaces.
\vspace{-1.2em}

\subsection{Experimental Setup}
\vspace{-0.8em}
We pretrain \method from a full-body motion tracking teacher trained on AMASS~\citep{mahmood2019amass}. 
The training split contains 11314 motion clips covering locomotion, turning, recovery, and diverse whole body motions. 

\vspace{-0.8em}
\begin{table}[t]
\centering
\caption{
Downstream action interfaces evaluated in our experiments. All latent interfaces use the same pretrained and frozen MotionPyramid. The first latent interface directly reuses the base per-step latent action decoder.
}
\begin{tabular}{llll}
\toprule
Interface & Decision rate & Policy output & Frozen decoder path \\
\midrule
Raw action & 30 Hz & \(z^{(0)}_t\) & none \\
\(z^{(1)}\) & 30 Hz & \(\Delta z^{(1)}_t\) & \(z^{(1)} \rightarrow z^{(0)}\) \\
\(z^{(2)}\) & 6 Hz & \(\Delta z^{(2)}_b\) & \(z^{(2)} \rightarrow z^{(1)} \rightarrow z^{(0)}\) \\
\(z^{(3)}\) & 3 Hz & \(\Delta z^{(3)}_B\) & \(z^{(3)} \rightarrow z^{(2)} \rightarrow z^{(1)} \rightarrow z^{(0)}\) \\
Mixture of Interfaces & learned & \(g,\Delta z^{(g)}\) & selected path \\
Residual Interfaces & 30, 6, 3 Hz & \(\Delta z^{(1)},\Delta z^{(2)},\Delta z^{(3)}\) & all paths active \\
\bottomrule
\end{tabular}
\label{tab:interfaces}
\end{table}

\subsection{Learning Speed and Precision Across Pyramid Levels}

\begin{table}[t]
\centering
\caption{
Tracking evaluation on AMASS-Train. Succ is rollout success rate; \(E_{g\text{-}\mathrm{mpjpe}}\) is global mean per-joint position error in millimeters; \(E_{g\text{-}\mathrm{rot}}\) and \(E_{l\text{-}\mathrm{rot}}\) are global and local rotation errors in degrees; \(J_{\mathrm{norm}}\) is a normalized joint-motion smoothness cost, where lower values indicate smoother motion.
}
\label{tab:amass_train_tracking_z_scratch}
\setlength{\tabcolsep}{5.5pt}
\renewcommand{\arraystretch}{1.08}
\begin{tabular}{lccccc}
\toprule
\multicolumn{6}{c}{\textbf{AMASS-Train}} \\
\midrule
Method
& Succ $\uparrow$
& $E_{g\text{-}\mathrm{mpjpe}}$ $\downarrow$
& $E_{g\text{-}\mathrm{rot}}$ $\downarrow$
& $E_{l\text{-}\mathrm{rot}}$ $\downarrow$
& $J_{\mathrm{norm}}$ $\downarrow$ \\
& & mm & deg & deg & \\
\midrule
scratch & 94.3\% & \textbf{58.4} & \textbf{13.6} & \textbf{10.8} & 675.3 \\
$z_1$   & \textbf{96.1\%} & 68.6 & 17.5 & 15.2 & 499.4 \\
$z_2$   & 92.0\% & 90.5 & 19.4 & 15.7 & 350.1 \\
$z_3$   & 81.0\% & 181.0 & 23.8 & 16.5 & \textbf{229.2} \\
\bottomrule
\end{tabular}
\end{table}

We first ask how control resolution affects learning both during distillation and during downstream task optimization. 
All comparisons in this section use the same humanoid, simulator, motion tracking teacher, and pretrained hierarchy design. 
For distillation, the only factor that changes is the pyramid level being trained. 
For downstream RL, all decoders and priors are frozen, and all policies use the same task reward, observation space, policy architecture, optimizer, initial state distribution, and training budget. 
This isolates the effect of the action interface itself.

\begin{figure}[t]
\centering

\begin{subfigure}[t]{0.48\linewidth}
    \centering
    \includegraphics[width=\linewidth]{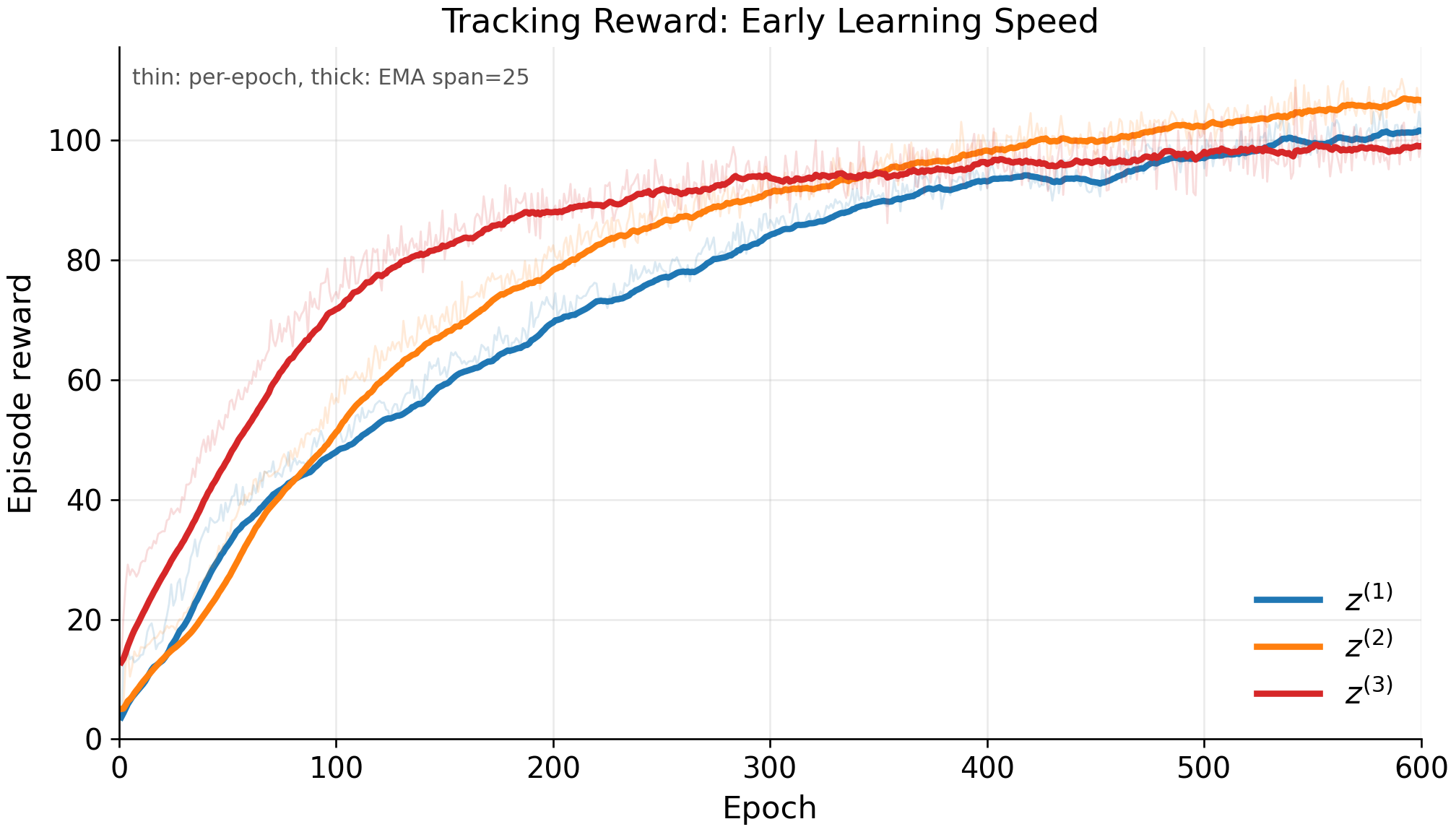}
    \caption{Early distillation learning}
    \label{fig:main_results_early_distill}
\end{subfigure}
\hfill
\begin{subfigure}[t]{0.48\linewidth}
    \centering
    \includegraphics[width=\linewidth]{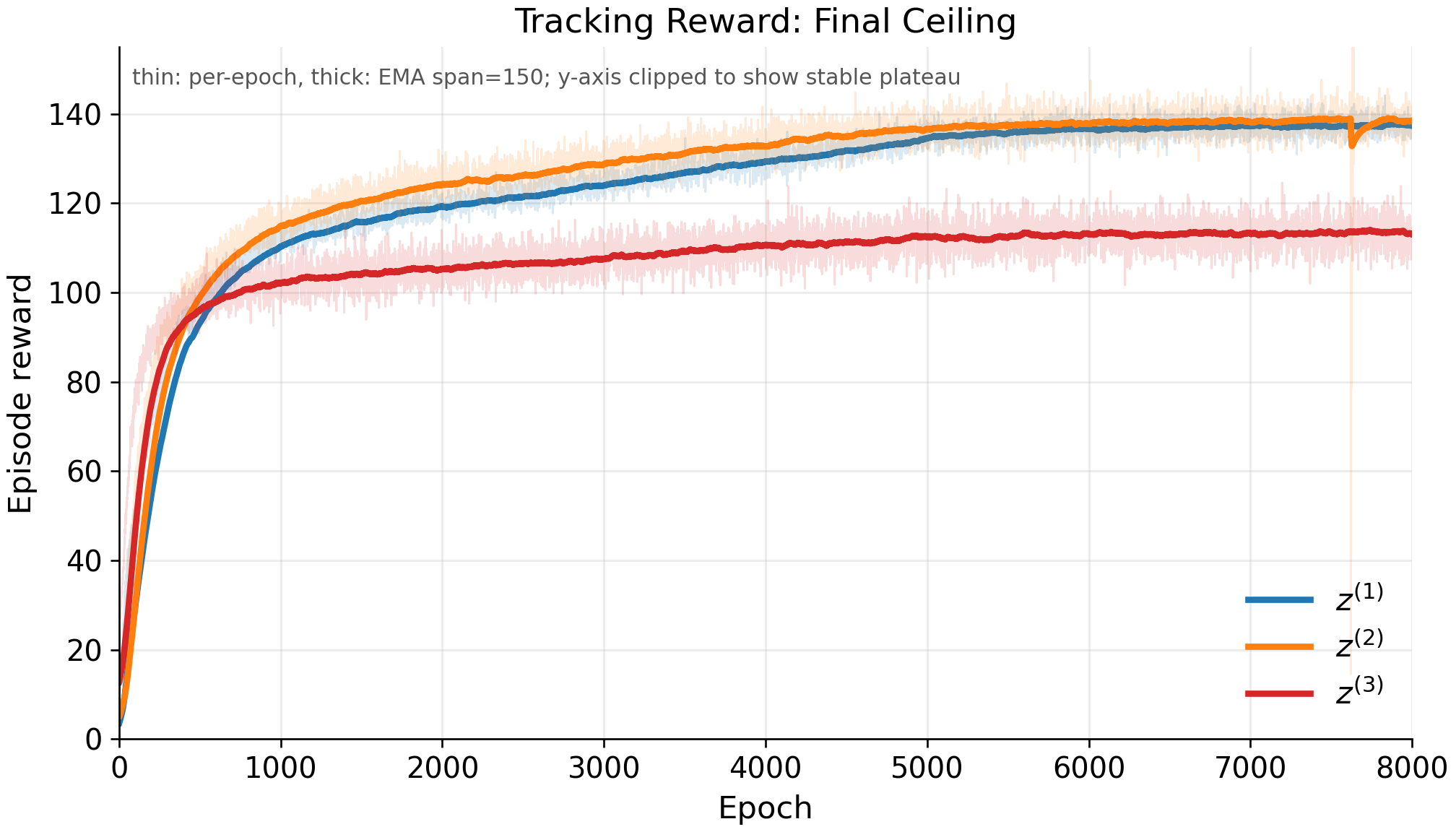}
    \caption{Final distillation ceiling}
    \label{fig:main_results_final_distill}
\end{subfigure}

\vspace{0.3em}

\begin{subfigure}[t]{\linewidth}
    \centering
    \includegraphics[width=\linewidth]{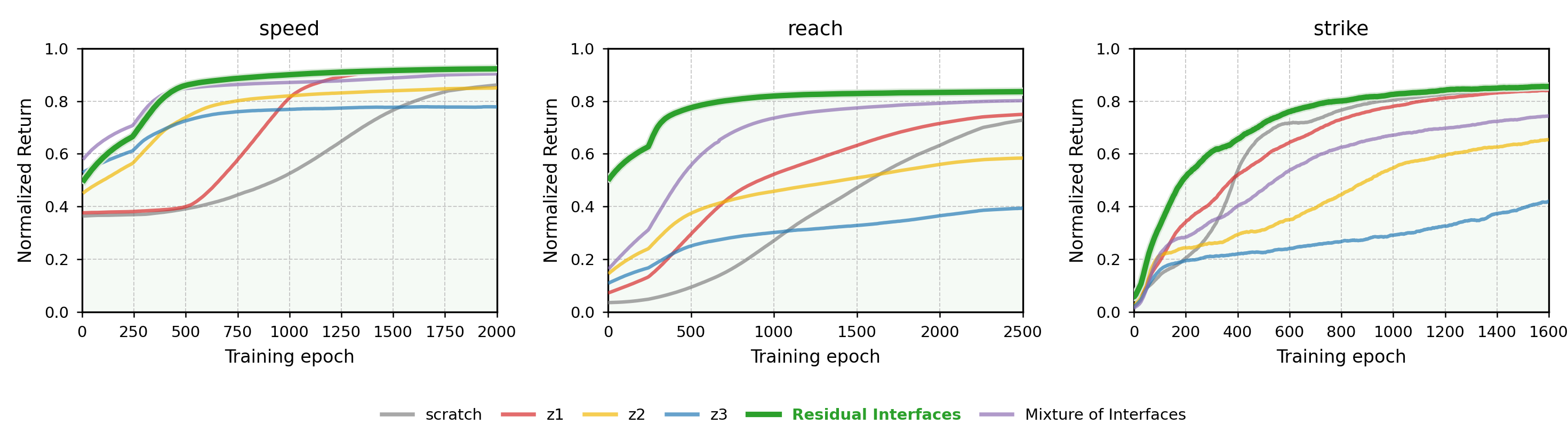}
    \caption{Adaptive downstream control on speed, reach, and strike}
    \label{fig:main_results_downstream}
\end{subfigure}

\caption{
\textbf{MotionPyramid improves both action representation learning and downstream control.}
Top: during recursive distillation, higher levels learn faster at early stages while lower levels preserve a stronger final control ceiling.
Bottom: for downstream reinforcement learning, we compare fixed MotionPyramid interfaces, Mixture of Interfaces, and Residual Interfaces on speed, reach, and strike.
Mixture of Interfaces selects one temporal interface online, while Residual Interfaces use a nested residual cascade that keeps all levels active at their native update rates.
Residual Interfaces achieve the strongest overall downstream performance, showing that multi level residual control combines the exploration benefits of coarse temporal abstraction with the precision of frame level correction.
}
\label{fig:main_results}
\vspace{-0.8em}
\end{figure}

Figure~\ref{fig:main_results}(a,b) evaluates the pretrained interfaces before
downstream RL. Temporal compression improves early optimization: higher levels
rise quickly because one latent explains a longer motion segment. The final
tracking ceiling shows the opposite. The fine and moderate interfaces
retain stronger tracking reward after long training, while the coarsest
interface converges to a lower plateau. This indicates that compression improves
early optimization but can discard fine tracking information.

Figure~\ref{fig:main_results}(c) evaluates the frozen interfaces on downstream
tasks. The tasks stress different aspects of humanoid control: speed emphasizes
stable locomotion, reach requires target alignment, and strike requires contact
timing and end effector precision. Across tasks, fixed interfaces form a
control frontier. The per step latent interface provides strong final
controllability, the moderate temporal interface often improves early learning
while remaining competitive, and the coarsest interface is more limited on
precision demanding tasks.

\name reaches strong final performance while retaining much of the
early learning behavior of temporally compressed interfaces. This suggests that
the pyramid levels provide complementary downstream action interfaces.

\begin{figure}[!t]
\centering
\includegraphics[width=\linewidth]{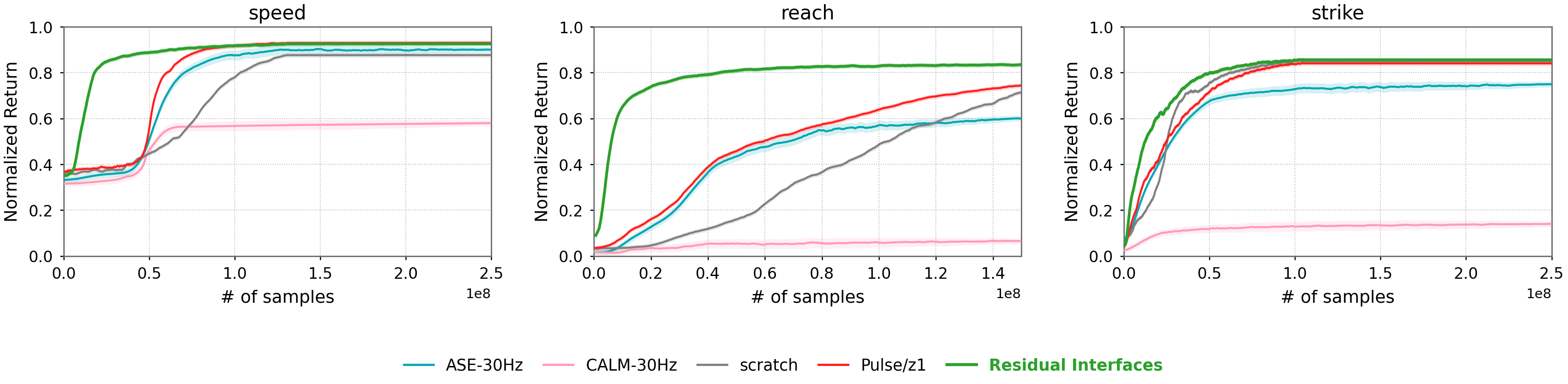}
\caption{
\textbf{Comparison against 30 Hz latent baselines.} We compare Residual Interfaces with scratch training and 30 Hz latent action baselines on speed, reach, and strike.}
\vspace{-1em}
\label{fig:ablation}
\end{figure}

Table~\ref{tab:amass_train_tracking_z_scratch} isolates the tracking behavior of our fixed interfaces, while Figure~\ref{fig:ablation} compares downstream control against 30 Hz latent baselines, including ASE-30Hz, CALM-30Hz, and the base per-step latent interface \(z^{(1)}\).

\vspace{-0.7em}
\subsection{Motion Naturalness from Sampling and Traversal}
\vspace{-0.7em}

\label{sec:sampling_traversal}

\begin{table}[t]
\centering
\small
\begin{tabular}{lccccc}
\toprule
Latent & Path Dist. (m) $\uparrow$ & Speed (m/s) $\uparrow$ & Action $\Delta$ $\downarrow$ & Min Root H. (m) $\uparrow$ & Fall: Low-Root Clips $\downarrow$ \\
\midrule
$z^{(1)}$ & $4.32 \pm 1.62$ & $0.433$ & $0.377$ & $0.343$ & $158/256$ \\
$z^{(2)}$ & $1.69 \pm 1.52$ & $0.170$ & $0.098$ & $0.822$ & $13/256$ \\
$z^{(3)}$ & $1.67 \pm 1.44$ & $0.168$ & $0.085$ & $0.867$ & $2/256$ \\
\bottomrule
\end{tabular}
\vspace{+0.4em}
\caption{
\textbf{Random rollouts under prior sampling for each latent layer.} Each clip lasts 10 seconds, and statistics are computed over 256 clips per layer. Low-root clips count rollouts whose minimum root height falls below 0.15 m, used as a collapse proxy because environment-level height termination is disabled. Lower-level latents produce larger XY path distances and higher action variation, while deeper latents produce smoother, more upright, and more stable rollouts.
}
\label{tab:latent_random_rollout_stats}
\end{table}

\begin{figure}[t]
\centering
\vspace{-2em}
\includegraphics[width=\linewidth]{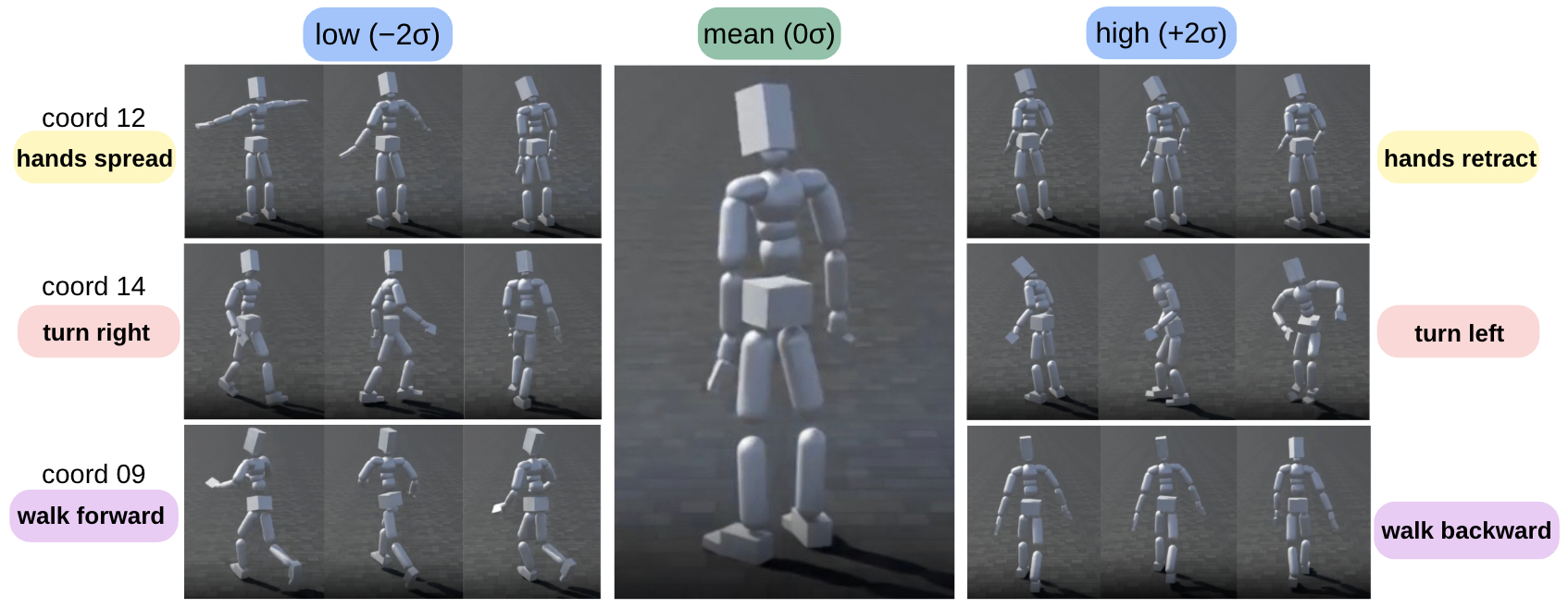}
\caption{
\textbf{Representative latent traversals at \(z^{(3)}\).}
For each row, we fix the proprioceptive state and all latent coordinates except one coordinate, then vary the selected coordinate from low \((-2\sigma)\) to the prior mean and high \((+2\sigma)\) in normalized prior units.
Since one \(z^{(3)}\) decision unfolds over \(H_3\) simulator steps, each cell visualizes a short rollout snippet rather than a single rendered frame.
The three rows show representative coordinates that induce noticeable and structured changes in decoded behavior.
}
\vspace{-1.5em}
\label{fig:z3_latent_traversal}
\end{figure}

We next evaluate whether the learned hierarchy improves motion regularity under unguided sampling and whether the learned action space contains continuous control directions.
For random sampling, we sample latents from the learned proprioceptive prior and decode them through the frozen hierarchy without any downstream task reward.
Table~\ref{tab:latent_random_rollout_stats} reports 10-second rollout statistics for each latent layer.

Random sampling reveals a clear effect of temporal abstraction.
Samples from \(z^{(1)}\) remain expressive and cover larger XY path distances, but they also exhibit higher action variation and lower minimum root height.
Samples from \(z^{(2)}\) and \(z^{(3)}\) produce shorter and smoother rollouts with more upright body configurations.
This pattern is consistent with the role of the pyramid: higher levels encode coarser motor programs, while lower levels retain finer moment to moment flexibility.

Latent traversal provides a complementary view.
Figure~\ref{fig:z3_latent_traversal} shows representative traversals at \(z^{(3)}\).
Because each \(z^{(3)}\) decision is held over \(H_3\) simulator steps, we visualize short rollout snippets rather than isolated frames.
By varying one coordinate while holding the current state and all other latent coordinates fixed, we observe noticeable and structured changes in decoded motion.
These traversals suggest that the learned representation contains continuous control directions, while also reflecting the smoother and more conservative behavior of higher-level temporal interfaces.

Overall, these results suggest that \method does more than compress action dimensionality.
It organizes motion into representations that support regular movement patterns and editable control directions at different temporal scales.
\vspace{-0.7em}
\subsection{Multi-Level Motion Editing and Reuse}
\label{sec:motion_editing}
\vspace{-0.7em}

\begin{figure}[t]
\centering
\includegraphics[width=\linewidth]{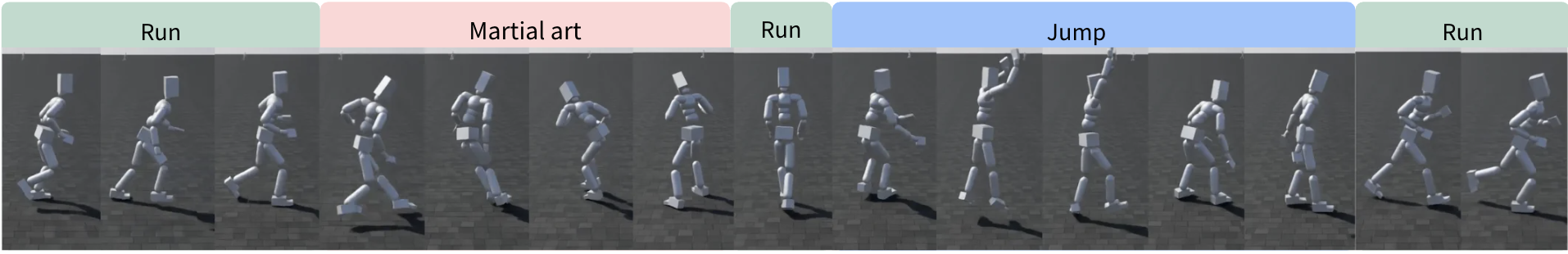}
\caption{
\textbf{Skill transition probe using the frozen MotionPyramid hierarchy.}
The rollout transitions between running, martial arts motion, running, jumping, and running again.
The sequence illustrates that the learned hierarchy can move between distinct motion modes while preserving physically stable whole body behavior.
}
\vspace{-1em}
\label{fig:transition_probe}
\end{figure}

We then examine whether the hierarchy exposes reusable control handles for editing and reusing motion.
Figure~\ref{fig:teaser}(c)(d) presents interpolation and composition demos, while Figure~\ref{fig:transition_probe} shows a dedicated transition probe.
These experiments are intended as representation probes.
They illustrate what kinds of motion structure are captured by the pyramid and how that structure can be manipulated across levels.

Interpolation shows that the hierarchy preserves smooth control directions across levels.
Varying a latent coefficient produces gradual changes in the decoded motion, indicating that the representation supports continuous motion editing.

Skill transition reveals a second property: a high-level latent can define a coarse motion context, while lower levels remain editable and can redirect local behavior within that context.
For example, holding a coarse latent while modifying intermediate or low-level latents produces smooth transition behavior without discarding the overall motion structure.
Figure~\ref{fig:transition_probe} visualizes this effect in a single rollout that transitions between running, martial arts motion, running, jumping, and running again.

Composition provides a third view of reuse.
Combining latents associated with different behaviors can produce fused motion patterns that inherit properties of both sources.
We treat this result as a qualitative probe rather than a full composition benchmark, but it supports the broader claim that the hierarchy captures reusable motion structure.
Taken together, these examples show that Motion Pyramid provides a set of fixed action abstractions for downstream reinforcement learning and also forms an editable multi-level motion representation.
\vspace{-0.7em}
\subsection{Adaptive Interfaces Resolve the Learning Speed and Control Precision Tradeoff}

Fixed MotionPyramid levels expose complementary downstream control regimes. The base per-step latent interface \(z^{(1)}\) acts at every simulator step and provides high frequency feedback, local recovery, and precise target alignment. Coarser interfaces reduce policy decision frequency and constrain exploration to temporally coherent motion segments, which can improve early learning and motion regularity. We evaluate two adaptive controllers that reuse the same frozen hierarchy.

\paragraph{Mixture of Interfaces.}
Mixture of Interfaces gives the downstream policy a discrete choice among the three pretrained interfaces. At each decision boundary, the policy selects a level \(g\in\{1,2,3\}\), predicts one residual latent at that level, and executes the corresponding frozen decoder path for its native duration. This controller is useful as an interpretable adaptive baseline because the selected horizon directly reveals where the policy spends feedback bandwidth.

\begin{figure}[t]
\centering
\includegraphics[width=\linewidth]{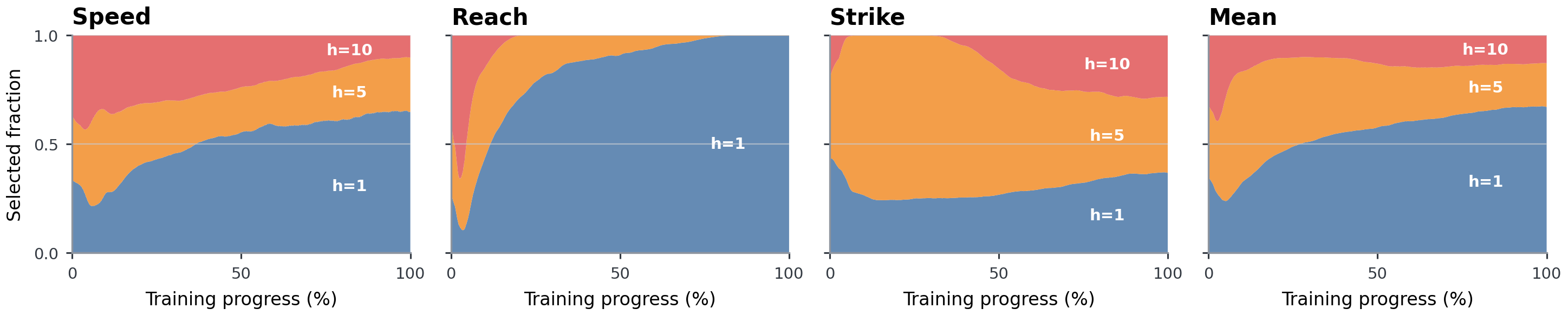}
\caption{
\textbf{Selection of Layers in Mixture of Interfaces.} We plot the fraction of selected horizons over downstream training for speed, reach, strike, and their mean.}
\vspace{-1em}
\label{fig:selection}
\end{figure}

Figure~\ref{fig:selection} analyzes the horizon choices made by Mixture of Interfaces. The selector does not collapse to a single temporal scale. Reaching gradually shifts toward \(H=1\), consistent with its need for accurate target alignment and local correction. Speed uses a mixture of horizons, while strike retains substantial use of \(H=5\) and \(H=10\), indicating that temporally extended motor programs remain useful for locomotion, ballistic movement, and contact preparation. These patterns show that MoI allocates decision bandwidth according to task demand.

\paragraph{Residual Interfaces.}
Residual Interfaces use a nested residual cascade. The controller maintains active latent commands at all hierarchy levels: \(z^{(3)}\) is refreshed every ten simulator steps, \(z^{(2)}\) every five steps, and \(z^{(1)}\) every simulator step. The high level latent provides a slow motion context, the middle level residual refines short motion segments, and the first level residual performs frame level correction through the frozen base decoder.

Overall, the comparison shows that adaptive multi level residual control adds value beyond single resolution latent action interfaces and scratch reinforcement learning.

%% file: chaps/related_work.tex
\vspace{-0.8em}
\section{Related Work}
\vspace{-0.8em}
\paragraph{Reusable motion latent spaces for control.}
A broad line of work learns reusable motor spaces for physically simulated characters.
Early approaches study hierarchical or compositional motor primitives that can be reused across tasks
\citep{wayne2014hierarchical,peng2019mcp,merel2018npmp,hasenclever2020comic,bohez2022imitate}.
Recent physics-based character controllers learn latent action spaces from adversarial objectives, variational models, discrete priors, multi-task action representations, or large-scale distillation from motion tracking teachers
\citep{peng2022ase,tessler2023calm,yao2022controlvae,won2022physicsvae,zhu2023ncp,luo2024universal,hua2023simple}.
These methods show that motion data can provide compact and reusable control spaces for downstream reinforcement learning.
Most of them expose a single downstream action interface, such as a per step latent action, a fixed horizon skill code, or a learned prior.
MotionPyramid extends this line by exposing the same pretrained motion structure as a multi-level family of action interfaces, allowing downstream policies to trade fine feedback control against coarser motion programs.
\vspace{-0.8em}
\paragraph{Kinematic motion priors and learned dynamics.}
Kinematic motion priors learn compact pose or transition models directly from motion data.
Motion VAE and HuMoR use variational sequence models to produce reusable human motion priors in pose space
\citep{ling2020mvae,rempe2021humor}.
Recent learned dynamics models such as Neural Motion Simulator predict future physical states from current observations and actions, and use accurate predicted rollouts for skill acquisition in imagined environments
\citep{hao2025mosim}.
These works provide complementary views of motion reuse at the pose or dynamics level.
MotionPyramid focuses on the action interface level, where pretrained decoders and priors are frozen and reused by closed loop downstream controllers.
\vspace{-0.8em}
\paragraph{Policy distillation and closed loop supervision.}
Our pretraining setup is also related to policy distillation, dataset aggregation, and kickstarting, where a student policy is shaped by one or more teacher policies
\citep{rusu2015policy,ross2011reduction,schmitt2018kickstarting}.
MotionPyramid uses a motion tracking teacher as supervision for latent action decoders.
After pretraining, the learned hierarchy is frozen, so downstream reinforcement learning operates through a reusable interface rather than relearning low-level motor coordination from task reward alone.
\vspace{-0.8em}
\paragraph{Structured and interpretable latent representations.}
Broader work on structured latent representations studies how learned spaces expose factors, transformations, and hierarchical structure.
CNN and transformer visualization methods probe representations through activation patterns, ablations, dictionary factors, and semantic structure
\citep{zeiler2014visualizing,yun2021transformer}.
Sparse Manifold Transform and RG Flow show that latent spaces can organize transformations, interpolation, scale separation, and semantic editing
\citep{chen2018sparse,hu2022rgflow}.
These works motivate our traversal and multi-level editing probes, where latent variables are evaluated as control handles for humanoid motion.
\vspace{-0.8em}
\paragraph{Hierarchical RL and temporal abstraction.}
Classical HRL formalizes reusable temporally extended decisions through feudal control, options, hierarchical machines, MAXQ, and skill chaining
\citep{vezhnevets2017feudal,sutton1999between,parr1997reinforcement,dietterich2000hierarchical,bacon2017option}.
Deep HRL learns multi timescale policies, subgoal interfaces, or latent skill priors for downstream control
\citep{vezhnevets2017feudal,nachum2018data,levy2019learning,nachum2019near,hausman2018learning,eysenbach2019diversity,pertsch2021accelerating}.
\textit{MotionPyramid} shares this temporal abstraction view and applies it to a frozen motion conditioned action interface for physics based humanoid control.

%% file: chaps/discussion.tex
\vspace{-0.7em}\section{Discussion and Limitations}
\vspace{-0.7em}

\method studies hierarchical motion representations as reusable residual interfaces between high-level decision making and low-level humanoid control. Our results suggest that motion data can be pretrained into a pyramid organized by temporal abstraction and controllability. Low-level latents preserve rapid feedback, contact correction, and target precision, while higher-level latents decode smoother and more temporally coherent motor programs. Fixed levels reveal a speed--precision tradeoff: coarse interfaces improve early learning and motion regularity by constraining exploration to structured motion segments, whereas fine interfaces retain the precision needed for alignment, timing, and recovery. Thus, the action interface is a representation choice that shapes downstream optimization, exploration, and the motion structure reused from pretraining.

The residual-interface view further enables adaptive use of the hierarchy. Downstream policies act through the frozen pyramid by predicting residual latents around pretrained motion priors rather than producing motor commands directly. When the policy can select the interface level online, it can route through coarse interfaces for temporally extended behavior and through fine interfaces for local correction, analogous in spirit to residual and skip pathways in deep architectures but applied to temporal control interfaces. Representation probes also show that traversal, interpolation, transition, and qualitative composition expose editable control handles across temporal scales, suggesting that the pyramid captures reusable motion structure beyond the downstream tasks.

Several limitations remain. First, downstream evaluation is limited to three simulated humanoid tasks, covering locomotion, target alignment, and contact timing but not the full range of dexterous, long-horizon, or real-world robot behavior. Second, the main downstream curves omit seed error bars due to computational cost, so statistical variability remains to be quantified. Third, the hierarchy is manually specified with $H_1=1$, $H_2=5$, and $H_3=10$; learning the number of levels, temporal horizons, and residual routing rules is an important direction for future work. Finally, our interpolation, transition, and composition results are qualitative probes. Developing quantitative benchmarks for multi-level motion editing and compositional reuse would provide a stronger test of hierarchical motion representations.

%% file: chaps/references.tex
\bibliographystyle{plainnat}
\bibliography{refs}

%% file: chaps/appendix.tex
\clearpage
\section{Method Details}
\label{app:method}

This appendix provides additional details for the construction of the recursive action interfaces and for downstream reinforcement learning with temporally extended latent decisions. We use \(z^{(1)}, z^{(2)}, z^{(3)}\) to denote the three learned action interfaces. The superscript indicates the temporal scale of the interface: \(z^{(1)}\) produces one simulator-step action, \(z^{(2)}\) unfolds over \(H_2=5\) simulator steps, and \(z^{(3)}\) unfolds over \(H_3=10\) simulator steps. We set \(H_1=1\).

\subsection{Environment and Teacher Policy}
\label{app:env_teacher}

We consider a humanoid control environment with physical state \(s_t\), policy observation \(o_t\), and low-level action \(a_t \in \mathcal{A}\). In our implementation, \(a_t\) corresponds to the low-level control command consumed by the simulator, such as target joint positions or PD targets. The environment evolves as
\[
s_{t+1} \sim P(s_{t+1} \mid s_t, a_t),
\]
and downstream tasks provide a scalar reward \(r_t = r(s_t, a_t, s_{t+1})\).

We first train or obtain a teacher policy \(\pi_T\) on the original low-level action space. The teacher is used only to construct the reusable action interfaces. After interface training, the teacher is discarded for downstream reinforcement learning. Downstream policies never output teacher actions directly; instead, they output latent interface decisions \(z^{(\ell)}\), which are decoded into low-level controls by the frozen interface.

\subsection{Recursive Action Interfaces}
\label{app:recursive_interface}

Each interface \(\mathcal{I}^{(\ell)}\) consists of a decoder \(D^{(\ell)}_{\phi}\). The base decoder maps a one-step latent decision into a low-level action:
\[
a_t = D^{(1)}_{\phi}(o_t, z_t^{(1)}).
\]
Higher-indexed interfaces are temporally extended. A latent \(z_t^{(\ell)}\) with \(\ell>1\) is selected at a decision time \(t\), held as the coarse latent context for \(H_\ell\) simulator steps, and recursively unfolded into lower-indexed latents. For \(j=0,\ldots,H_\ell/H_{\ell-1}-1\), we write
\[
z_{t+jH_{\ell-1}}^{(\ell-1)}
=
D^{(\ell)}_{\phi}
\left(
o_{t+jH_{\ell-1}},
z_t^{(\ell)},
j
\right).
\]
The generated \(z^{(\ell-1)}\) is then executed by the lower interface \(\mathcal{I}^{(\ell-1)}\). This recursion continues until the base decoder produces simulator actions.

For example, a \(z^{(2)}\) decision unfolds into five \(z^{(1)}\) latents, each of which produces one low-level action. A \(z^{(3)}\) decision unfolds into two \(z^{(2)}\) latents, and each \(z^{(2)}\) further unfolds into five \(z^{(1)}\) latents. Thus \(z^{(3)}\) controls a ten-step segment through recursive latent unfolding:
\[
z^{(3)}
\rightarrow
z^{(2)}
\rightarrow
z^{(1)}
\rightarrow
a.
\]

Importantly, temporally extended interfaces are not open-loop action sequences. Although \(z_t^{(\ell)}\) is selected only once every \(H_\ell\) simulator steps, the decoder receives the current observation during unfolding. Therefore, the decoded lower-level latents and actions remain closed-loop with respect to the humanoid state.

\subsection{Closed-Loop Interface Distillation}
\label{app:distillation}

We train the interfaces bottom-up. First, we train \(D^{(1)}_{\phi}\) to reproduce teacher behavior through the one-step latent interface. We then freeze \(D^{(1)}_{\phi}\) and train \(D^{(2)}_{\phi}\) to generate \(z^{(1)}\) sequences that reproduce teacher behavior over \(H_2\) steps. Finally, we freeze the lower decoders and train \(D^{(3)}_{\phi}\) to generate lower-indexed latent decisions over \(H_3\) steps.

For each level \(\ell\), we introduce a tracking latent controller \(\mu_{\psi}^{(\ell)}\) during distillation:
\[
z_t^{(\ell)} \sim \mu_{\psi}^{(\ell)}(\cdot \mid o_t, c_t),
\]
where \(c_t\) denotes the tracking command or reference information used during interface training. The tracking controller is used only to learn the decoder. Downstream policies later replace \(\mu_{\psi}^{(\ell)}\) with task-specific policies.

The distillation objective is evaluated in closed loop. Starting from an initial state, the latent controller selects \(z^{(\ell)}\), the decoder recursively unfolds it into low-level actions, and the simulator advances under the decoded actions. We then compare the resulting rollout against the teacher or reference rollout. A generic loss for level \(\ell\) is
\[
\begin{aligned}
\mathcal{L}_{\mathrm{distill}}^{(\ell)}
&=
\mathbb{E}
\Bigg[
\sum_{i=0}^{H_\ell-1}
\Big(
\lambda_a
\left\|
\hat a_{t+i}
-
a^T_{t+i}
\right\|_2^2
+
\lambda_q
d_q
\left(
\hat q_{t+i},
q^T_{t+i}
\right)
\\
&\qquad\qquad\qquad
+
\lambda_v
\left\|
\hat v_{t+i}
-
v^T_{t+i}
\right\|_2^2
+
\lambda_e
\left\|
\hat e_{t+i}
-
e^T_{t+i}
\right\|_2^2
\Big)
\Bigg]
+
\lambda_z \mathcal{R}_z.
\end{aligned}
\]
Here \(\hat a,\hat q,\hat v,\hat e\) denote the decoded action, body pose, body velocity, and end-effector features from the interface rollout, while \(a^T,q^T,v^T,e^T\) denote the corresponding teacher or reference quantities. The term \(\mathcal{R}_z\) regularizes the latent space, for example through latent norm, KL, or smoothness penalties depending on the implementation.

The key point is that the loss is computed on states visited by the learned interface, not only on teacher-forced states. This closed-loop training reduces distribution shift between interface distillation and downstream usage.

\subsection{Downstream Control with Frozen Interfaces}
\label{app:downstream}

For downstream reinforcement learning, all interface decoders are frozen. A downstream policy only learns to select latent decisions. For a fixed interface level \(\ell\), the policy is
\[
z_n^{(\ell)} \sim \pi_{\theta}^{(\ell)}(\cdot \mid o_{t_n}, g_{t_n}),
\]
where \(g_{t_n}\) denotes the downstream task observation or command, and \(t_n\) is the simulator time at the \(n\)-th policy decision. The selected latent is executed for \(H_\ell\) simulator steps through the frozen recursive decoder. The next policy decision occurs at
\[
t_{n+1} = t_n + H_\ell,
\]
unless the episode terminates earlier.

Thus, the downstream policy acts at a lower decision frequency for higher-indexed interfaces. The environment still advances at the original simulator frequency, and rewards are accumulated at every simulator step.

\subsection{Semi-Markov Formulation}
\label{app:smdp}

Temporally extended latent decisions induce a semi-Markov decision process over policy decision times. Let \(t_n\) be the simulator time of the \(n\)-th latent decision. The policy selects a latent decision \(z_n\) that is executed for \(h_n\) simulator steps. For fixed-interface policies, \(h_n=H_\ell\); for Mixture of Interfaces policies, \(h_n\) depends on the selected interface level.

The discounted reward accumulated during one latent decision is
\[
\bar r_n
=
\sum_{i=0}^{m_n-1}
\gamma^i r_{t_n+i},
\]
where \(\gamma\) is the per-simulator-step discount factor and \(m_n \le h_n\) is the actual number of simulator steps executed before the next decision or episode termination.

The corresponding SMDP discount is
\[
\bar \gamma_n
=
\begin{cases}
\gamma^{m_n}, & \text{if the episode does not terminate during the segment},\\
0, & \text{otherwise}.
\end{cases}
\]
The high-level transition stored in the replay buffer is therefore
\[
\left(
o_{t_n},
z_n,
\bar r_n,
\bar \gamma_n,
o_{t_{n+1}}
\right).
\]
For \(z^{(1)}\), we have \(h_n=1\), \(\bar r_n=r_{t_n}\), and \(\bar\gamma_n=\gamma\), recovering the standard MDP formulation.

\subsection{SMDP Generalized Advantage Estimation}
\label{app:smdp_gae}

We train downstream policies with PPO using generalized advantage estimation adapted to the SMDP decision times. Let \(V_{\theta}(o_{t_n})\) be the value estimate at the \(n\)-th latent decision. The SMDP temporal-difference residual is
\[
\delta_n
=
\bar r_n
+
\bar \gamma_n
V_{\theta}(o_{t_{n+1}})
-
V_{\theta}(o_{t_n}).
\]
The generalized advantage estimate is computed recursively over decision indices:
\[
\hat A_n
=
\delta_n
+
\bar \gamma_n \lambda \hat A_{n+1},
\]
where \(\lambda\) is the GAE parameter. Equivalently,
\[
\hat A_n
=
\sum_{k=0}^{\infty}
\left(
\prod_{j=0}^{k-1}
\bar \gamma_{n+j}\lambda
\right)
\delta_{n+k}.
\]
The value regression target is
\[
\hat V_n = \hat A_n + V_{\theta}(o_{t_n}).
\]

For fixed-level interfaces, this amounts to computing GAE over every \(H_\ell\)-th simulator state while using the discounted reward accumulated over the skipped simulator steps. For Mixture of Interfaces policies, the same equations apply with variable \(h_n\) and variable \(\bar\gamma_n\).

\subsection{PPO Objective for Temporally Extended Decisions}
\label{app:smdp_ppo}

For a fixed interface level \(\ell\), the PPO likelihood ratio is computed only at latent decision times:
\[
\rho_n(\theta)
=
\frac{
\pi_{\theta}^{(\ell)}
\left(
z_n^{(\ell)}
\mid
o_{t_n}, g_{t_n}
\right)
}{
\pi_{\theta_{\mathrm{old}}}^{(\ell)}
\left(
z_n^{(\ell)}
\mid
o_{t_n}, g_{t_n}
\right)
}.
\]
The clipped policy objective is
\[
\mathcal{L}_{\pi}
=
-
\mathbb{E}_n
\left[
\min
\left(
\rho_n(\theta)\hat A_n,
\mathrm{clip}
\left(
\rho_n(\theta),
1-\epsilon,
1+\epsilon
\right)
\hat A_n
\right)
\right].
\]
The value loss is
\[
\mathcal{L}_{V}
=
\mathbb{E}_n
\left[
\left(
V_{\theta}(o_{t_n})-\hat V_n
\right)^2
\right].
\]
The full PPO objective is
\[
\mathcal{L}_{\mathrm{PPO}}
=
\mathcal{L}_{\pi}
+
c_V \mathcal{L}_{V}
-
c_H \mathcal{H}
\left[
\pi_{\theta}^{(\ell)}
\right],
\]
where \(c_V\) and \(c_H\) are the value and entropy coefficients.

Rollouts are collected for the same number of simulator steps across all interfaces. Higher-indexed interfaces therefore produce fewer policy decision samples per rollout, but each decision sample contains a longer SMDP transition with the corresponding accumulated reward and discount.

\subsection{Mixture of Interfaces Interface Selection}
\label{app:Mixture of Interfaces}

For Mixture of Interfaces policies, the action at decision time \(t_n\) consists of both an interface level and a latent decision:
\[
\ell_n \sim p_{\theta}(\ell \mid o_{t_n}, g_{t_n}),
\qquad
z_n^{(\ell_n)}
\sim
\pi_{\theta}^{(\ell_n)}
(\cdot \mid o_{t_n}, g_{t_n}).
\]
The selected latent is decoded by the corresponding frozen interface \(\mathcal{I}^{(\ell_n)}\) and executed for
\[
h_n = H_{\ell_n}
\]
simulator steps. The next policy decision occurs after the selected interface finishes execution or when the episode terminates.

The log probability used by PPO is
\[
\log \pi_{\theta}
\left(
\ell_n,
z_n^{(\ell_n)}
\mid
o_{t_n}, g_{t_n}
\right)
=
\log p_{\theta}
\left(
\ell_n
\mid
o_{t_n}, g_{t_n}
\right)
+
\log
\pi_{\theta}^{(\ell_n)}
\left(
z_n^{(\ell_n)}
\mid
o_{t_n}, g_{t_n}
\right).
\]
The PPO ratio is therefore
\[
\rho_n(\theta)
=
\exp
\left[
\log \pi_{\theta}
\left(
\ell_n,
z_n^{(\ell_n)}
\mid
o_{t_n}, g_{t_n}
\right)
-
\log \pi_{\theta_{\mathrm{old}}}
\left(
\ell_n,
z_n^{(\ell_n)}
\mid
o_{t_n}, g_{t_n}
\right)
\right].
\]
All advantage estimates are computed using the SMDP GAE equations from Appendix~\ref{app:smdp_gae}, with the duration \(h_n\) determined by the selected level.

For budgeted Mixture of Interfaces experiments, we can additionally penalize decision cost by modifying the accumulated SMDP reward:
\[
\bar r_n^{\mathrm{budget}}
=
\sum_{i=0}^{m_n-1}
\gamma^i r_{t_n+i}
-
\eta c_{\ell_n},
\]
where \(c_{\ell_n}\) measures the cost of selecting interface level \(\ell_n\), and \(\eta\) controls the strength of the budget penalty. Unless otherwise stated, evaluation reports the original unpenalized task reward.

\subsection{Decision Bandwidth}
\label{app:bandwidth}

We measure the decision bandwidth of an interface by the number of policy-level latent decisions required per simulator step. For a fixed interface \(\ell\), the decision frequency is
\[
f_{\mathrm{dec}}^{(\ell)}
=
\frac{f_{\mathrm{sim}}}{H_\ell},
\]
where \(f_{\mathrm{sim}}\) is the simulator control frequency. When latent dimensionality is included, the normalized latent bandwidth is
\[
B^{(\ell)}
=
\frac{d_\ell}{H_\ell},
\]
where \(d_\ell\) is the dimensionality of \(z^{(\ell)}\).

For Mixture of Interfaces policies, the average decision frequency over an episode is
\[
\bar f_{\mathrm{dec}}
=
f_{\mathrm{sim}}
\frac{N}{\sum_{n=1}^{N} h_n},
\]
where \(N\) is the number of policy decisions in the episode. The corresponding average latent bandwidth is
\[
\bar B
=
\frac{
\sum_{n=1}^{N} d_{\ell_n}
}{
\sum_{n=1}^{N} h_n
}.
\]
These quantities allow us to compare fixed interfaces and Mixture of Interfaces policies in terms of downstream return, smoothness, and policy decision cost.

\subsection{Implementation of Rollout Storage}
\label{app:rollout_storage}

During downstream PPO training, the simulator is stepped at the original control frequency for all methods. The policy is queried only at interface decision times. For each latent decision, we accumulate the discounted simulator-step rewards until the next decision:
\[
\bar r_n
=
r_{t_n}
+
\gamma r_{t_n+1}
+
\cdots
+
\gamma^{m_n-1} r_{t_n+m_n-1}.
\]
We then store a single SMDP transition in the PPO buffer. Intermediate simulator states are not treated as policy decision states, although they affect the accumulated reward, termination condition, and next observation.

This implementation ensures that temporally extended interfaces are trained under the same underlying environment objective as one-step policies. The only difference is the temporal resolution at which the downstream policy is allowed to make decisions.

\subsection{Multiscale Editing Probe}
\label{app:multiscale_editing}

For qualitative analysis, we also use the recursive structure of the interface to edit latent decisions at different temporal scales. Given two latent rollouts, we can interpolate a latent at level \(m\) as
\[
\tilde z^{(m)}
=
(1-\alpha) z_A^{(m)}
+
\alpha z_B^{(m)},
\qquad
\alpha \in [0,1].
\]
When \(m=3\), interpolation changes the coarse ten-step motion context. When \(m=2\), it modifies a shorter segment inside the \(z^{(3)}\) context. When \(m=1\), it edits frame-level details such as contact, balance, or end-effector pose.

This probe is used only to visualize the structure of the learned interface. It does not change the training objective. The purpose is to show that the recursive decoder exposes editable control handles at multiple temporal scales: a coarse latent can define a longer motion context, while lower-indexed latents can still induce local transitions or refinements during unfolding.

\section{Implementation Details and Hyperparameters}
\label{app:hyperparams}

All policies are trained with PPO using Adam. Unless otherwise specified, we use
$\gamma=0.99$, GAE parameter $\lambda=0.95$, PPO clipping threshold $0.2$,
gradient clipping at $50$, and observation normalization with clipping range
$[-5,5]$. Actor and critic networks use SiLU activations. For downstream RL
experiments, we use 32 rollout steps per environment and 6 PPO epochs per
update. The actor and critic are separate MLPs with hidden sizes
$(2048,1024,512)$. The policy log standard deviation is fixed during training.

\begin{table}[h]
\centering
\small
\resizebox{\linewidth}{!}{%
\begin{tabular}{lcccccc}
\toprule
Policy & Interface & Action dim. & Horizon & Actor/Critic MLP & Log std. & LR \\
\midrule
$z^{(0)}$ & raw PD target & 69 & 1 & $(2048,1024,512)$ & $-2.9$ & $1{\times}10^{-5}$ / $2{\times}10^{-5}$ \\
$z^{(1)}$ & first latent decoder & 32 & 1 & $(2048,1024,512)$ & $-1.0$ & $2{\times}10^{-5}$ / $2{\times}10^{-5}$ \\
$z^{(2)}$ & second latent decoder & 32 & 5 & $(2048,1024,512)$ & $-1.0$ & $2{\times}10^{-5}$ / $2{\times}10^{-5}$ \\
$z^{(3)}$ & third latent decoder & 32 & 10 & $(2048,1024,512)$ & $-1.0$ & $2{\times}10^{-5}$ / $2{\times}10^{-5}$ \\
MoI & learned interface selection & 32 & $\{1,5,10\}$ & $(2048,1024,512)$ & $-1.0$ & $2{\times}10^{-5}$ / $2{\times}10^{-5}$ \\
\bottomrule
\end{tabular}
}
\caption{
\textbf{Downstream policy architectures.} MoI denotes Mixture of Interfaces. For MoI,
the selector chooses among $z^{(1)}$, $z^{(2)}$, and $z^{(3)}$ interfaces with
horizons $1$, $5$, and $10$, respectively. All learned interfaces use
32-dimensional latent actions. The raw-action baseline uses a robot action scale
of $0.3$.
}
\label{tab:downstream_hyperparams}
\end{table}

The Mixture of Interfaces policy adds a categorical horizon/interface selector
on top of the shared actor observation. The selector is a two-layer MLP with
hidden sizes $(512,256)$ and outputs logits over $\{1,5,10\}$. We train the
selector and latent policy jointly with SMDP actor updates: actor losses are
applied only at option boundaries, while the critic remains a per-step value
function. The SMDP advantage uses GAE over option durations. For MoI downstream
runs, we use an option-boundary cost of $0.02$, applied only after the task
reward exceeds $0.85$, with a ramp factor of $0.1$.

\begin{table}[h]
\centering
\small
\resizebox{\linewidth}{!}{%
\begin{tabular}{lcccccc}
\toprule
Decoder & Latent dim. & Output dim. & Plan horizon & Goal mode & Goal horizon & Distill LR \\
\midrule
$z^{(1)}$ decoder & 32 & 69 & 1 & single-frame tracking & 1 & $3{\times}10^{-5}$ \\
$z^{(2)}$ decoder & 32 & 32 & 5 & future single & 5 & $3{\times}10^{-5}$ \\
$z^{(3)}$ decoder & 32 & 32 & 10 & future dense & 5 & $3{\times}10^{-5}$ \\
\bottomrule
\end{tabular}
}
\caption{
\textbf{Latent decoder training hyperparameters.} All decoders are VAE-style policies
with a learned prior, SiLU activations, observation normalization, and MLP
hidden sizes $(1536,1024,512)$ for task encoders and $(2048,1024,512)$ for
latent-action heads, except the first decoder, whose action head uses
$(3096,2048,1024)$. The KL coefficient is annealed from $0.01$ to $0.001$ for
$z^{(1)}$, from $0.03$ to $0.01$ for $z^{(2)}$, and from $0.04$ to $0.015$ for
$z^{(3)}$. We also use an AR(1) smoothness penalty with weight $0.005$ and
$\phi=0.99$.
}
\label{tab:decoder_hyperparams}
\end{table}

For speed, reach, and strike, downstream policies are trained with 512 parallel
environments. The default downstream batch size is 1024, except for the raw
speed baseline where we use batch size 4096 for improved PPO stability. Speed
targets are sampled in $[0,5]$ m/s and resampled every 100--200 steps. Reach
uses the right hand as the end-effector with target changes every 50--100 steps.
Strike uses right-hand contact bodies and a dynamic target object. All reported
downstream results use the frozen pretrained decoders; only the downstream
policy and, for MoI, the interface selector are updated.

\section{Compute Resources}
\label{app:compute}

All experiments were run on a local GPU workstation. The machine contains
8 NVIDIA RTX 4090 GPUs with 24GB memory each, two AMD EPYC 7763 CPUs
(128 physical cores / 256 threads total). Experiments
were implemented in PyTorch 2.7 with CUDA 12.8 and Isaac Lab simulation.

\begin{table}[h]
\centering
\small
\begin{tabular}{lccc}
\toprule
Experiment type & GPUs per run & Parallel envs & Notes \\
\midrule
Raw-action downstream RL & 4 & 512 & Direct 69D PD-action baseline \\
$z^{(1)}$ downstream RL & 4 & 512 & Frozen $z^{(1)}$ decoder \\
$z^{(2)}$ downstream RL & 4 & 512 & Frozen $z^{(2)}$ decoder, SMDP updates \\
$z^{(3)}$ downstream RL & 4 & 512 & Frozen $z^{(3)}$ decoder, SMDP updates \\
MoI downstream RL & 4 & 512 & Learns interface selector over $\{1,5,10\}$ \\
$z^{(1)}$ decoder distillation & up to 8 & 2048 & Distilled from PHC teacher \\
$z^{(2)}$ decoder distillation & up to 8 & 2048 & Distilled from $z^{(1)}$ teacher \\
$z^{(3)}$ decoder distillation & up to 8 & 2048 & Distilled from $z^{(2)}$ teacher \\
\bottomrule
\end{tabular}
\caption{
\textbf{Compute setup used for the main experiments.} Downstream results use frozen
decoders and train only the task policy, while decoder distillation trains the
latent interface itself. MoI denotes Mixture of Interfaces.
}
\label{tab:compute_resources}
\end{table}

For downstream PPO experiments, each run used 32 simulation steps per rollout
and 6 PPO epochs per update. The batch size was 1024 for most downstream runs;
the raw-action speed baseline used batch size 4096 for improved PPO stability.
Decoder distillation used 32-step rollouts, one optimization epoch per update,
and Adam with learning rate $3\times 10^{-5}$. We checkpointed long-running
training jobs every 1000 epochs and used the latest available checkpoint for
figures unless otherwise specified.
\subsection{Assets and Licenses}

We use AMASS motion data for motion tracking supervision and cite the original
dataset. Access to AMASS follows the dataset license and terms of use. The
simulation environment is based on Isaac Lab, and learning components are
implemented in PyTorch. Any released code will include license information and
instructions for obtaining external assets from their original sources.